# Particle Swarm Optimization with Velocity Restriction and Evolutionary Parameters Selection for Scheduling Problem

P.V. Matrenin, V.G. Sekaev
Department of Automated Control System
Novosibirsk State Technical University
Novosibirsk, Russia
pavel.matrenin@gmail.com
sekaev@mail.ru

*Abstract*—The article presents a study of the Particle Swarm optimization method for scheduling problem. To improve the method's performance a restriction of particles' velocity and an evolutionary meta-optimization were realized. The approach proposed uses the Genetic algorithms for selection of the parameters of Particle Swarm optimization. Experiments were carried out on test tasks of the job-shop scheduling problem. This research proves the applicability of the approach and shows the importance of tuning the behavioral parameters of the swarm intelligence methods to achieve a high performance.

## I. INTRODUCTION

Combinatorial optimization is one of the most important areas of discrete mathematic. Thousands of industrial practice tasks can be formulated as combinatorial optimization problems. Real-life problems often belong to NP-hard problems and have a high dimensionality. For solving such problems stochastic optimization methods are commonly used, as these methods allow receiving good solutions in a reasonable time.

The stochastic optimization method known as Particle Swarm Optimization (PSO) was designed by J. Kennedy and R. Eberhart and it was based on a bird flock's behavior [1]. PSO works by maintaining a swarm of particles that move around in the multidimensional search space influenced by the improvements discovered by the other particles. The method can be employed easily for many optimization problems. However, according to the No Free Lunch theorem [2] and a number of special researches such as [3], [4], the PSO method cannot always work well, and requires tuning the parameter for each optimization problem or each type of problem. At present the most authoritative researcher engaged in the selection of the parameters is M. Pedersen, but he only deals with continuous optimization problems.

In this study the job-shop scheduling problem is considered as a test problem. It is a NP-hard combinatorial problem; in addition, drawing even one valid schedule is a nontrivial task. Therefore, the number of algorithm's iterations are limited, especially for operational management tasks that require the rapid generation of schedules in real time.

The behavioral parameters of PSO significantly influence contraction of PSO to a single point (some solution of an optimization problem). Controlling convergence and contraction is the important subtask for researchers using PSO to practice. A way to improve the performance of the PSO method by the meta-optimization and the controlling contraction was proposed and analyzed in this study.

The article is organized as follows. Section II gives a brief overview of the PSO method. Section III describes the job-shop scheduling problem, which was used as a benchmark. Section IV presents the PSO variance proposed and the method for adaptation used for tuning the parameters. Section V gives experimental evidence, and the conclusion describes the results.

## II. PARTICLE SWARM OPTIMIZATION

The task of finding a minimum (maximum) of some function $f(X)$ is solved, where $X$ is the vector of the variable parameters which may possess values from some space $D = \{d_{1\,min}, d_{1\,max}, \ldots, d_{s\,min}, d_{s\,max}\}$, where $s$ is the dimensionality of the search space. Every particle is denoted by $X$ coordinates from $D$ and by the value of the optimizing function $f(X)$. Vector $X$ denotes the position of a particle, and vector $V$ is particle's velocity. Initial values $X$ and $V$ are random, $X \in D$ ($x_i \in [d_{i\,min}, d_{i\,max}]$, $i = 1, \ldots, n$). Velocity $V$ is limited by vector $V_{max}$. Then all particles' $X$ and $V$ are updated iteratively according to the following formulas [1], [3]:

$$V \leftarrow V\omega + \alpha_1(P - X)R_1 + \alpha_2(G - X)R_2, \quad (1)$$

$$\text{if } (V > V_{max}), \text{ then } V \leftarrow V_{max}$$
$$\text{else if } (V < -V_{max}), \text{ then } V \leftarrow -V_{max},$$

$$X \leftarrow X + V,$$

where *P* is particle's previous best position, which is found by the particle among the decision space,

*G* is the previous best position of the whole swarm position, which is found by the swarm among the decision space,

$R_1$ and $R_2$ are random vectors, elements of which are uniformly distributed within the space (0, 1). Vectors $R_1$ and $R_2$ have the same lengths as *X*, *V*, *P*, *G* and *D*.

The behavioral parameters $\alpha_1$ and $\alpha_2$ define weights of particle's attraction to points *P* and *G*. Parameter ω characterizes inertia of the particles, and parameter-vectors $V_{max}$ limits particle's velocity *V*.

## III. SCHEDULING PROBLEM

The job-shop scheduling problem is among the hardest combinatorial optimization problems. The problem consists in scheduling a finite set of jobs (tasks, problems) *N* on a finite set of machines (devices, performers, etc.) *M* with the objective of minimizing the makespan. The makespan is the total length of the schedule that is the maximum of completion times needed for processing all jobs. The process of servicing each job includes several stages (operations), i.e. each job has a special order through the machines during an uninterrupted time interval. One machine can process at most one operation at a time. A schedule can be described as an allocation of the operations to time intervals on the machines. The problem is to find the shortest (quickest) schedule. The job-shop is a model that can be considered as a basis for many real-life scheduling problems; hence, it is very important problem.

This problem is NP-hard generally [5]. It is known that small size instances of the problem can be solved with a reduced computational time by exact algorithms, such as brand-and-bounds. However, for large instances, only heuristic (approximate) algorithms achieve satisfactory results. Let [6]:

- $N = \{1,\ldots, n\}$ is the set of jobs.
- $M = \{1,\ldots,m\}$ is the set of machines.
- $V = \{0,1,\ldots, j+1\}$ denote the set of the operations, 0 and *j+1* are fictions operations: start and finish.
- *A* be the set of pair of operations constrained by the precedence relations.
- $V_j$ be the set of operations to be performed by the machine *j*.
- $E_k \subset V_k \times V_k$ be the set of pair of operations to be performed on the machine *k* and which therefore have to be sequenced.
- $p_v$ and $t_v$ denote the known processing time and the unknown start time of the operation *v*.

Given this assumption the job-shop problem can be considered as:

*minimize* $t_{j+1}$ $\quad t_j - t_i = p_i$, $\quad (i, j) \in A$

*subject to*

$$t_j - t_i \geq p_i \ \& \ t_i - t_j \geq p_j, \ (i, j) \in E_k, k \in M \quad (2)$$

Any feasible solution of the problem (2) is called schedule.

## IV. VELOCITY RESTRICTION AND PARAMETER SELECTION

### A. Velocity restriction

Vector $V_{max}$ equals $(D_{max} - D_{min})$ generally and it means that any particle can cross the entire search space during a single step. The high speed of the particles allows the algorithm to converge quickly to some local extremum. For example, studies [1], [3], [7] touch upon tuning the parameters of PSO and do not consider velocity-boundaries as a variable parameter, so $V_{max}$ is a constant. However, when particles move over long distances, they can miss extremes that are more effective (in terms of the optimality criterion). As a result, the particles can contract to some local inefficient extremum.

There are PSO variants making velocity-boundaries narrower for each step, for example [8]. However, in the later steps of the algorithm particles are not accelerated to high speeds as a general rule, since the summands (*P* – *X*) and (*G* – *X*) in equation (1) decrease when the particles contract around a local extremum. In addition, this method does not solve the problem of missing good extremums during the first iterations of the algorithm, when the velocities are high.

The research presented by this paper proposes to limit the velocity (velocity restriction) at once and to not change it during execution of the algorithm. In this case

$$V_{max} = \beta \ (D_{max} - D_{min}),$$

where β is a new behavioral parameter of the PSO method. The PSO set β equals 1, but for our approach β much less than 1.0. The approach reduces the risk of missing global or near-global extremums. At the same time it is simple to implement as the original method in contrast to the above mentioned approach of dynamically lowering the velocity-boundaries.

On the other hand, this study experimentally found that low velocity reduces the effectiveness of the algorithm in the later iterations, because the particles cease to find new solutions, contracting around some extreme found. To increase the diversification properties (ability to find new solutions) the coefficient β was set not less than 0.01. Thus, the method has four parameters whose values must choose to obtain a high efficiency.

The problem of the best performing parameters selection can be considered as an optimization task in its own right, and solved by another optimization method (known as meta-optimization) [3]. Some optimization method is used as an overlaying meta-optimizer searching the parameters of other optimization method to provide a high performance.

### B. Adaptation

Our previous research [9] suggests applying the Genetic algorithm (GA) as a meta-optimizer for tuning the parameters of Ant Colony Optimization to solve the job-shop scheduling problem. GA is chosen as the analog of evolutionary selection. The same method of meta-optimizing was used in this study, for the PSO with the velocity restriction. This approach can be described as follows: the optimization problem is solved by swarm intelligence algorithm, and GA is used as an overlaying meta-optimizer, which provides the parameters of swarm intelligence algorithm, in this case, it is PSO. The parameters $\{\alpha_1, \alpha_2, \omega, \beta\}$ are used as genes so one set of parameters is one chromosome. The solution quality obtained by the usage of these parameters is taken as the value of the fitness function. The best parameters are selected for the next iterations, going through crossover and mutation. After crossover and mutation, a new population of parameters appears. Thus, the adaptation is achieved for every current task. The method obtained has been named the Adaptive Swarm Intelligence algorithm, Adaptive PSO (APSO) in this case.

Recommendation parameters for the classic general-purpose PSO parameters is found in [10], where $\alpha_1 = \alpha_2 = 1.49445$, $\omega = 0.729$, $\beta = 1.0$. The research [7] gives a number of good parameters for various optimization scenarios. For problem considered the closest match of parameters is {-0.2746, 4.8976, -0.3488, 1.0}. It is these sets (Kennedy [10] {1.49445, 1.49445, 0.729, 1.0}, Pedersen [7] {-0.2746, 4.8976, -0.3488, 1.0}) were used in experiments. Let's denote these sets of the parameters as PSO K. and PSO P. respectively.

*C. Adaptive algorithm*

Since PSO is a stochastic method, it gives different results for each optimization run. Hence a result of one optimization run is not a reliable criterion for the parameters optimization. A simple way of overcoming this difficulty is to perform a number of runs ($k$ runs), and then to take the average of the results obtained. The approach's outline is shown below.

*1)* Initialize the population, i.e. initialize each chromosome with random values of corresponding behavior parameters of the PSO algorithm.

*2)* Until a termination criterion is met, repeat the following.

   *a)* For each chromosome in the current population:

   *i)* Solve the optimization problem by PSO with behavior parameters of this chromosome $k$ times saving values of a criteria.

   *ii)* Write an average value of the criteria as fitness.

   *iii)* If this value is better than the best-known fitness, then save the value as the best-known fitness and save the corresponding parameters as the best-known parameters.

   *b)* Perform standard steps of GA, such as selection, crossover and mutation (this research used mutation probability of 10% and one-point crossover).

*3)* Now the best-known parameters are the parameters desired for the PSO method for solving the optimization problem.

This approach is easy to implement and very flexible. The main disadvantage is the large computation time, but, on the other hand, the approach can be performed in parallel easily and efficiently, because it is possible to do parallelization at the level of GA [9].

V. EXPERIMENTAL RESULTS

The algorithms were implemented in C++ language on a 2.4 GHz Intel CPU i7 and was tested on a number of job-shop problem instances denoted as LA01-LA21 of various sizes [11] provided by OR-Library [12]. The sizes of instances are from 10×5 to 15×10 ($n \times m$).

The original PSO with parameters values recommended by researches [7] (PSO P.) and [10] (PSO K.) and presented in Section II has been run 100 times for all instances, and an average and the best solutions have been found. The APSO with the velocity restriction has been run only one time with 100 iterations and 50 chromosomes of GA. The number of optimization runs for finding the average result ($k$) was 10. Both variants used 50 particles and 100 iterations of PSO.

For training the APSO instances LA02, LA18 and LA20 were selected arbitrarily. The set of parameters obtained from training was $\{\alpha_1 = 1.76428, \alpha_2 = 1.38203, \omega = 0.730135, \beta = 0.280868\}$. This set is used for all instances to the same conditions that PSO P. and PSO K. Table I and II show the results (average and best) obtained by these experiments and well known results from the literature [6].

TABLE I. THE AVERAGE RESULTS BY PSO WITH PARAMETERS COMPARED

| Instance | PSO K. | PSO P. | APSO | Well known |
|---|---|---|---|---|
| LA01 | 670.42 | 669.52 | 690.8 | 666 |
| LA02 | 687.19 | 684.83 | 693.35 | 655 |
| LA03 | 627.11 | 626.43 | 541.49 | 597 |
| LA04 | 610.25 | 610.73 | 613.79 | 590 |
| LA05 | 593 | 593 | 593 | 593 |
| LA06 | 926 | 926 | 926.03 | 926 |
| LA07 | 891.61 | 891.08 | 902.8 | 890 |
| LA08 | 867.25 | 865.65 | 882.09 | 863 |
| LA09 | 951.11 | 951.13 | 955.34 | 951 |
| LA10 | 958 | 958 | 958 | 958 |
| LA11 | 1222 | 1222 | 1222.05 | 1222 |
| LA12 | 1039.01 | 1039.03 | 1043.59 | 1039 |
| LA13 | 1150.09 | 1150.01 | 1161.59 | 1150 |
| LA14 | 1292 | 1292 | 1292 | 1292 |
| LA15 | 1226.79 | 1228.57 | 1283.31 | 1207 |
| LA16 | 992.2 | 983.12 | 1011.11 | 945 |
| LA17 | 806.11 | 805.74 | 819.67 | 784 |
| LA18 | 892.8 | 889.16 | 930.19 | 848 |
| LA19 | 898.83 | 896.75 | 865 | 842 |
| LA20 | 944.51 | 950 | 907 | 902 |
| LA21 | 1175.04 | 1188 | 1128 | 1046 |

TABLE II. THE BEST RESULTS BY PSO WITH PARAMETERS COMPARED

| Instance | PSO K. | PSO P. | APSO | Well known |
|---|---|---|---|---|
| LA01 | 666 | 670 | 666 | 666 |
| LA02 | 658 | 663 | 658 | 655 |
| LA03 | 597 | 622 | 597 | 597 |
| LA04 | 590 | 594 | 590 | 590 |
| LA05 | 593 | 593 | 593 | 593 |
| LA06 | 926 | 926 | 926 | 926 |
| LA07 | 890 | 890 | 890 | 890 |
| LA08 | 863 | 863 | 863 | 863 |
| LA09 | 951 | 951 | 951 | 951 |
| LA10 | 958 | 958 | 958 | 958 |
| LA11 | 1222 | 1222 | 1222 | 1222 |
| LA12 | 1039 | 1039 | 1039 | 1039 |
| LA13 | 1150 | 1150 | 1150 | 1150 |
| LA14 | 1292 | 1292 | 1292 | 1292 |
| LA15 | 1207 | 1207 | 1207 | 1207 |
| LA16 | 979 | 982 | 956 | 945 |
| LA17 | 784 | 804 | 784 | 784 |
| LA18 | 859 | 896 | 859 | 848 |
| LA19 | 866 | 894 | 865 | 842 |
| LA20 | 911 | 950 | 907 | 902 |
| LA21 | 1123 | 1188 | 1128 | 1046 |

Let's suppose that the unit of the schedule's time for the test instances is 1 hour. Thus, the total deviation between the average and well-known results appeared as 455.32 hours of PSO with parameters by J. Kennedy and R. Eberhart, 723.38 hours for PSO with parameters by M. Pedersen and 433.83 hours for the Adaptive PSO. Thus, the APSO has proved to be 5% and 66% more efficient than two other comparable variants. The failure of PSO with parameters by M. Pedersen can be explained by the fact that the parameters recommended by Pedersen were obtained for continuous optimization problems, however, as the parameter of Kennedy and Eberhart.

The largest computation time of single solving one instance was about 0.48 seconds, the least time was 0.08 seconds (these values are the same for all three PSO variants, since the algorithms differ only by the parameters). The training process lasted 7.8 hours (without parallelization); the parallelization would reduce training time by several times depending on the number of cores.

## CONCLUSION

This research considers the Particle Swarm Optimization method for the job-shop scheduling problem in terms of the operational control, which results in a serious limitation of computing time, i.e. the number of algorithm's iterations. To improve the efficiency of PSO the velocity restriction was considered as an additional variable behavioral parameter of the method.

To achieve a high performance of this PSO variance the adaptation to problems' conditions was applied by tuning the behavioral parameters. The Genetic algorithm was used as a meta-optimizer for this adaptation.

Three job-shop scheduling problem instances were allocated among the number of test instances and the adaptation was carried out using these three training instances. The resulting values of the parameters were then used to solve a number of test job-shop instances.

The experiments showed the following.

*1)* The PSO method allows obtaining solutions of job-shop scheduling problems very close to the best known results. Moreover, the method is fast, since during the experiments the method used only 100 iterations.

*2)* Using inappropriate parameters can give poor results even if these parameters are very effective for problems of another class.

*3)* The restriction of the particles' velocity and the evolutionary adaptation of the parameters through Genetic algorithm has improved the performance of the PSO method significantly.

*4)* It was not found that the adaptation method is subject to overtuning since good results were shown not only for instances where the parameters selection was executed, but also for other instances.

*5)* The surprisingly high efficiency of parameters' values proposed by Kennedy and Eberhard [10] should be noted, taking into account that they have selected the values for completely different problems. Therefore, in cases where there is no opportunity for the selection of the parameters, it is reasonable to use these values ($\alpha_1 = \alpha_2 = 1.49445$, $\omega = 0.729$, $\beta = 1.0$.).

Areas for further work are investigating the evolutionary meta-optimizing, applying the adaptation to other swarm algorithms, and developing the parallel implementation. In addition, the dynamic adaptation to the real-time changes is an important question for further investigation.

REFERENCES

[1] J. Kennedy, R. Eberhart, "Particle Swarm Optimization", in proc. of IEEE International Conference on Neural Network, Piscataway, NJ, 1995, pp. 1942-1948.
[2] D.H. Wolpert, W.G. Macready, "No Free Lunch Theorems for Optimization", IEEE Transactions on Evolutionary Computation vol. 1, no. 1, 1997, pp.67-82.
[3] M. Pedersen, A. Chippereld, "Simplifying Particle Swarm Optimization", Applied Soft Computing, vol. 10, issue 2, 2010, pp. 618-628.
[4] Y. Shi, R.C. Eberhart, "Parameter selection in particle swarm optimization", in proc. of Evolutionary Programming VII (EP98), 1998, pp. 591- 600.

[5] M. Garey, D. Johnson, R. Sethy, "The complexity of flow shop and job shop scheduling", in Mathematics of Operations Research, no. 1, 1976, pp. 117-129.

[6] F. Pezzella, E. Merelli, "A tabu search method guided by shifting bottleneck for the job shop scheduling problem" in European Journal of Operational Research, no. 120, 2000, pp. 297-310.

[7] M. Pedersen, "Good Parameters for Particle Swarm Optimization", Hvass Laboratories Technical Report no. HL1001, 2010.

[8] H. Fan, "A modification to particle swarm optimization algorithm" in Engineering Computations: International Journal for Computer-Aided Engineering, no. 19, 2002, pp. 970-989.

[9] P.V. Matrenin, V.G. Sekaev, "Optimizing adaptive ant colony algorithm on the example of scheduling problem" in Programmnaya inzheneriya, no. 4, 2013, pp. 34-40. (П.В. Матренин, В.Г. Секаев, "Оптимизация адаптивного алгоритма муравьиной колонии на примере задачи календарного планирования", Программная инженерия).

[10] R.C. Eberhart, Y. Shi, "Particle swarm optimization: developments, applications and resources", in proc. of the Congress on Evolutionary Computation, vol. 1, 2001, pp. 81-86.

[11] S. Lawrence, "Supplement to "resource constrained project scheduling: an experimental investigation of heuristic scheduling techniques", tech. rep., GSIA, Carnegie Mellon University, October 1984.

[12] J.E. Beasley, "OR-Library: distributing test problems by electronic mail", in Journal of the Operational Research Society, no. 41(11), 1990, pp. 1069-1072.